\documentclass[11pt]{article}
\usepackage[marginratio=1:1,height=600pt,width=460pt,tmargin=100pt]{geometry}

\usepackage{graphicx}
\usepackage[utf8]{inputenc} 
\usepackage[T1]{fontenc}
\usepackage{fix-cm}
\usepackage[hidelinks, colorlinks=true,citecolor=blue]{hyperref} 
\usepackage{url}
\usepackage{booktabs}
\usepackage{amsfonts}
\usepackage{nicefrac}
\usepackage{xcolor}
\usepackage[normalem]{ulem}
\usepackage{authblk}
\usepackage{amsthm,amsmath,amsfonts,amssymb}
\usepackage{mathtools}
\usepackage{mathrsfs}
\usepackage{color}
\usepackage{natbib}
\usepackage{caption}
\usepackage{subcaption}
\usepackage{bm}
\usepackage{longtable}
\usepackage[flushleft]{threeparttable}
\usepackage[shortlabels]{enumitem}
\usepackage{wrapfig}
\setlist{leftmargin=6mm}
\usepackage{multirow}
\usepackage{titlesec}
\usepackage{setspace}
\usepackage{pifont} 

\usepackage{algorithm}
\usepackage{algorithmic}
\newlength\myindent
\newtheorem{conjecture*}{Conjecture}
\theoremstyle{remark}

\allowdisplaybreaks

\title{LLM-Guided Transportation Hub Capacity Planning\\ with Textual Business Inputs}

\author{Xiaoyue Liu}
\author{Zheng Dong}
\affil{{\small H. Milton Stewart School of Industrial and Systems Engineering, Georgia Institute of Technology}\vspace{2pt}}
\date{}

\begin{document}

\maketitle

\begin{abstract}

While traditional hub capacity planning models optimize effectively for quantitative inputs, they often fail to digest qualitative business context. 
We propose a novel framework where a large language model (LLM) agent iteratively proposes hub capacity decisions guided by natural-language business context descriptions.
The key mechanism is a chain-of-thought reasoning protocol: the LLM constructs a structured decision table that maps each contextual item to specific capacity adjustments based on the implied direction and magnitude of changes. The new capacity decision is then validated through a feedback loop with an optimization model, which provides routing-based performance metrics to guide the agent’s selection.
On a real-world 13-hub freight network in the southeastern US, our framework achieves a 2.8\% optimality gap relative to the hidden ground-truth, a significant improvement over the 11.0\% gap produced by the traditional optimization model without textual business inputs.
This demonstrates that LLMs can serve as a contextual bridge, integrating qualitative business insights into Operations Research workflows.
\end{abstract}



\section{Introduction}

Real-world hub capacity planning is rarely a purely quantitative exercise. While stochastic optimization provides a rigorous foundation for decision-making under uncertainty~\citep{birge2011introduction}, they face inherent limitations in digitizing qualitative information, including regulatory changes and labor market disruptions, that practitioners routinely consider when making real-world decisions. Traditional approaches either ignore such textual context or require manual translation into numerical parameters, a process that is both non-scalable and error-prone.

The emergence of Large Language Models (LLMs) as reasoning agents offers a potential solution to this bottleneck. LLMs excel at parsing unstructured natural-language descriptions and performing high-level qualitative assessments~\citep{bommasani2021opportunities}.
Recent work has begun exploring the synergy between LLMs and optimization. ~\cite{wang2025solid} coordinates an optimization agent and an LLM agent via dual prices for portfolio optimization. ~\cite{jovine2025listen} uses LLMs as zero-shot preference oracles for multi-objective selection. ~\cite{ahmaditeshnizi2024optimus} utilizes LLMs to translate natural language into mathematical model formulations. ~\cite{liu2024dellma} applies LLMs to general decision-making under uncertainty. 
However, a critical gap remains: how can the reasoning capabilities of LLMs be integrated with mathematical optimization to enhance decision-making where critical contextual information resides outside the model?

We propose a framework for hub capacity planning in transportation networks where: an LLM agent interprets natural-language business context to propose capacity levels, an optimization model provides routing evaluation as an oracle, and each iterative refinement loop computes hub utilization based on routing feedback to make keep/discard decisions.
Our contributions are threefold: 
\begin{enumerate}[itemsep=1.5pt, parsep=1.5pt, topsep=1pt]
    \item We propose a novel framework coupling LLM chain-of-thought reasoning with two-stage stochastic optimization for transportation hub capacity planning under qualitative uncertainty. 
    \item We design a structured Business Context Decision Table protocol that enables LLMs to systematically map textual business inputs to hub capacity adjustments.
    \item Our experimental validation on a real freight network demonstrates that the LLM agent can identify complex network structures and achieves a 2.80\% optimality gap against a hidden ground-truth model, effectively bridging quantitative optimization methods with critical textual information.
\end{enumerate}

\begin{figure*}[t]
\centering
\includegraphics[width=0.98\textwidth]{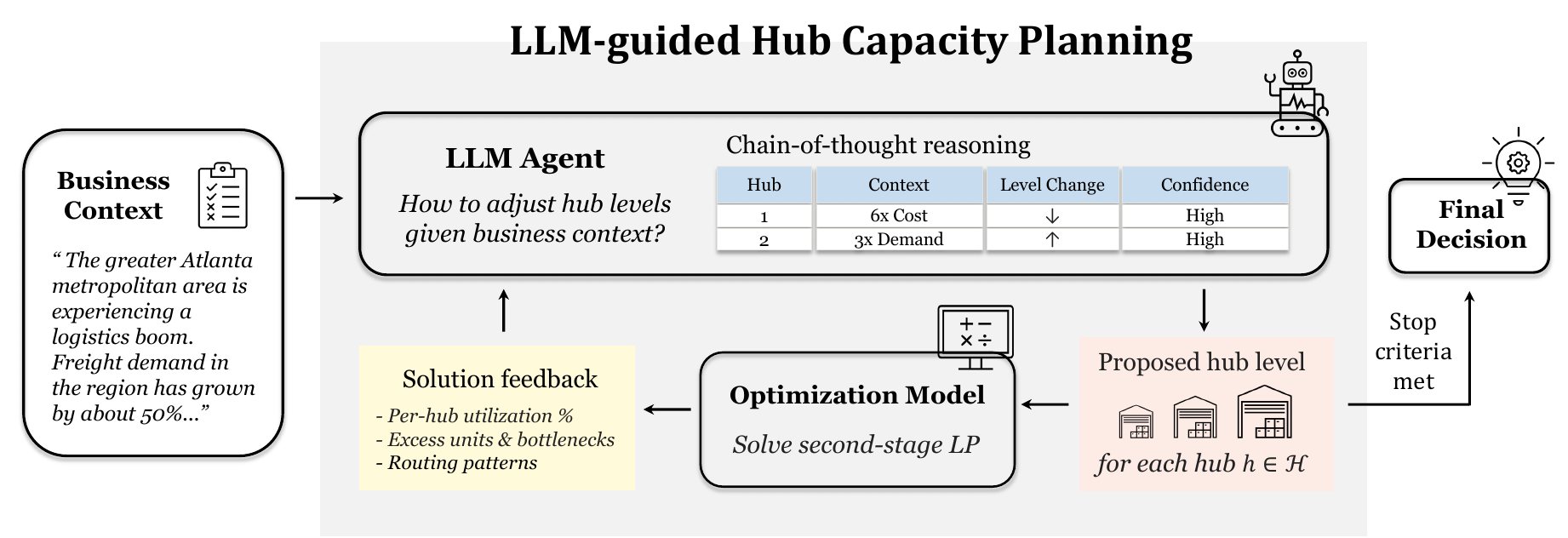}
\caption{Overview of the proposed LLM-guided hub capacity planning framework. The LLM agent receives textual business context and uses chain-of-thought reasoning to design hub levels using a structured decision table. The proposed solutions are evaluated by an optimization model that solves the second-stage routing problem and returns solution feedback. This feedback loop iterates until a stopping criterion is met, producing the final capacity decision.}
\label{fig:framework}
\end{figure*}

\section{Related Work}
\label{app:related}

Our work draws on three research streams: hub location under uncertainty, LLMs as reasoning agents, and frameworks coupling LLMs with optimization solvers.

Hub location problems are well-studied in network optimization~\citep{alumur2008network}, with stochastic variants using two-stage formulations where hub investments are committed before demand is realized~\citep{contreras2011stochastic, birge2011introduction}.
Recent extensions address balanced capacity utilization~\citep{hu2021stochastic, liu2025network}, hub disruption risks~\citep{liu2024multi}, dynamic resource allocation~\citep{liu2024dynamic, liu2026modular, liu2026two}, and robust profit-maximizing objectives~\citep{merakli2021robust}. 
In freight relay transportation specifically, hyperconnected hub networks enable long-haul shipments via short-haul drivers commuting between fixed-base hubs~\citep{liu2023logistics, li2024stochastic}.
Our work builds on this relay structure but introduces qualitative business context that modifies cost and demand parameters in ways existing stochastic formulations do not address.
The quality of two-stage solutions depends not only on the optimization model but also on how demand uncertainty is characterized prior to capacity commitment. While classical hub location models traditionally depend on stationary Poisson processes for scenario generation, more expressive non-stationary spatio-temporal point processes~\citep{zhu2022neural, dong2023spatiotemporal, dong2023non} could be substituted to capture spatial-temporal correlation in freight demand, with graph-structured variants modeling events over connected domains~\citep{dong2026deep} and conditional generative formulations enabling direct scenario sampling~\citep{dong2025conditional}.
Integrating these advanced event-level generative techniques ensures that the downstream hub-and-relay configurations remain both robust and highly adaptive to complex, shifting market conditions.

Chain-of-Thought prompting~\citep{wei2022chain} and Tree-of-Thoughts~\citep{yao2024tree} have enabled LLMs to tackle multi-step reasoning tasks.
LLMs have been explored as direct optimizers via iterative prompt refinement~\citep{yang2024large, dong2025greentea} and as hyper-heuristics~\citep{ye2024reevo}, though they consistently struggle with precise numerical computation at scale.
DeLLMa~\citep{liu2024dellma} proposes a structured framework for LLM decision-making under uncertainty but operates independently of mathematical programming solvers.
Recent surveys~\citep{huang2025orlm, ramamonjison2023nl4opt} provide broader overviews of LLM--OR interactions.

The most related work integrates LLMs with optimization solvers directly.
OptiMUS~\citep{ahmaditeshnizi2024optimus} and Chain-of-Experts~\citep{xiao2024chainofexperts} use LLMs to formulate or decompose optimization problems from natural language.
SOLID~\citep{wang2025solid} coordinates optimization and LLM agents via an ADMM-inspired mechanism with dual prices, demonstrating improved portfolio returns when combining quantitative precision with contextual understanding from financial news.
LISTEN~\citep{jovine2025listen} uses LLMs as zero-shot preference oracles for selecting from Pareto-optimal sets via utility refinement and tournament methods.
In supply chain planning, recent efforts integrate LLMs with network models for interactive explanation~\citep{venkatachalam2025integrating, aghaei2025potential}, though they focus on visualization rather than driving optimization decisions.

Our framework differs in three respects: (1)~we address stochastic network design with discrete capacity decisions, unlike the continuous portfolio weights in SOLID or selection tasks in LISTEN; (2)~the LLM reasons about qualitative context modifying a hidden cost model it never observes, creating an information asymmetry by design; and (3)~we deliberately separate routing feedback from cost feedback: while SOLID communicates dual prices, we instruct the agent to disregard the original model's reported costs because they do not reflect business context effects, preventing the optimizer from overriding correct context-driven decisions.

\section{Problem Description}

In freight transportation networks, carriers contract with hub providers to lease capacity at intermediate transfer points, enabling long-haul shipments to be carried by multiple short-haul drivers~\citep{li2023multi, liu2025dynamic}.
A central planning challenge is hub capacity planning: before demand is realized, the carrier must decide how much capacity to contract at each hub in the network.
Contracting too little leads to costly overflow penalties when demand exceeds capacity; contracting too much wastes resources on unused capacity.
The primary decision of this problem is a capacity level $l_h$ for each hub $h$ in the network, which represents the contracted weekly throughput at that location.
The traditional optimization approach formulates this as a two-stage stochastic program on a network $\mathcal{G} = (\mathcal{N}, \mathcal{A})$ with hub nodes $\mathcal{H} \subseteq \mathcal{N}$ and origin-destination (OD) pairs $\mathcal{K}$.
In the \emph{first stage}, the planner commits to capacity levels for all hubs before demand uncertainty is resolved.
In the \emph{second stage}, given the realized demand across scenario set $\Omega$, freight is routed through the network to minimize transportation and overflow penalty costs.
The objective minimizes total hub costs plus expected second-stage costs (see full formulation in Appendix~\ref{app:formulation}).

Traditional optimization methods use parameters estimated based on historical data and expert judgment. However, in practice, inputs feeding into this optimization are not static and such estimates may not reflect the latest conditions.
The up-to-date information about these changes comes in the form of textual business context: natural-language descriptions such as ``Hub in Savannah area is facing new EPA environmental compliance requirements; operating expenses have increased roughly ninefold'' (see Appendix~\ref{app:context} for the full set).
These descriptions convey the direction and approximate magnitude of input changes, but not exact numerical values.
Then, our research question is: \emph{Can an LLM agent, given only these textual business descriptions and routing feedback from an optimization model, produce capacity decisions that account for the latest qualitative context?}

\section{Methodology}

\paragraph{Overall Framework}
Our framework consists of three key components, as shown in Figure~\ref{fig:framework}:
\begin{itemize}[itemsep=3pt, parsep=3pt, topsep=3pt]
    \item \textbf{Business Context}: Natural-language descriptions of qualitative factors that affect hub costs, penalties, and demand. For example, it may include labor market disruptions, regulatory changes, new infrastructure investments, temporary demand surge and so on.
    \item \textbf{LLM Agent}: 1) Receives textual business contexts and optimization feedback from the last iteration; 2) Uses chain-of-thought reasoning to propose/modify a set of capacity level decisions $\{l_h\}_{h \in \mathcal{H}}$ for all hubs in the network; 3) Determines whether \emph{keep} or \emph{discard} the proposed hub level solution in the current iteration.
    \item \textbf{Optimization Model}: Evaluates each proposed first-stage capacity solution by solving the second-stage routing problem, and returns per-hub utilization percentages and excess units as feedback signals.
\end{itemize}
A critical design principle is that the optimization model serves as a routing evaluation only: its cost outputs are deliberately suppressed because they are computed from the original parameters, which do not reflect the latest business context changes.
Only routing-based metrics are returned to the LLM to assess how traffic distributes across hubs and identify capacity mismatches that need adjustment, guiding hub capacity adjustments through observed network behaviors. This design prevents the LLM agent from being misled by cost signals that contradict the qualitative context inputs.

\paragraph{Chain-of-Thought Reasoning}
The LLM does not propose capacity levels in a single step.
Instead, we construct a Business Context Decision Table through a chain-of-thought reasoning process.
For each hub affected by a business context item, the LLM reasons through four fields within a single structured response:
\begin{enumerate}[itemsep=3pt, parsep=3pt, topsep=3pt]
    \item \textbf{Context interpretation}: \emph{What does the business context description imply for this hub?} For instance, a cost increase suggests reducing capacity investment; a demand surge suggests more capacity needed; a penalty increase suggests less overflow tolerance.
    \item \textbf{Magnitude assessment}: \emph{How strong is the capacity change signal?} The LLM maps qualitative language to approximate scales (e.g., ``change significantly'' = very large, ``decrease by about half'' = large, ``roughly 20\% higher'' = moderate, ``slightly above normal'' = small).
    \item \textbf{Cost-benefit reasoning}: \emph{Is it cheaper to pay penalties or contract more capacity?} The LLM compares the effective hub cost at each capacity level against the expected penalty savings, using the context-implied cost structure.
    \item \textbf{Confidence assignment}: \emph{How reliable is this capacity adjustment?} Each capacity change decision is tagged as High (strong signal with clear direction), Medium (clear direction but uncertain magnitude), or Low (ambiguous or marginal effect).
\end{enumerate}
For hubs without any business context, the LLM right-sizes capacity based on observed routing patterns from the optimization feedback.
High-confidence decisions are held firm across multiple iterations; low-confidence decisions remain adjustable based on new routing evidence.

\paragraph{Iterative Refinement Loop}
The full LLM-guided hub capacity planning procedure is given in Algorithm~\ref{alg:loop}.
At each iteration, the optimization model evaluates the current capacity level solution and returns routing feedback. 
The LLM then makes the keep or discard decision on the proposed solution based on two criteria: 1) whether the proposed solution respects all high-confidence business context decisions, and 2) whether the resulting routing patterns are reasonable (no unexpected flow bottlenecks or wasted capacity allocation). 
Finally, the LLM updates the decision table with the new routing feedback and proposes the next capacity level candidate.
Crucially, this keep/discard decision does not use the optimization model's cost objective: those costs are biased when computed with the original parameters, and injecting the latest business context inputs can inherently changes the parameters. So the agent is instructed to disregard them and may keep a solution even when the base-model cost rises.
In the end, the loop terminates when a maximum iteration count is reached, the solution stabilizes, or the LLM agent indicates no further adjustments are needed.
The LLM prompt template and the keep/discard guidelines are detailed in Appendix~\ref{app:prompt}.

\begin{algorithm}[!t]
   \caption{LLM-Guided Hub Capacity Planning}
   \label{alg:loop}
\begin{algorithmic}[1]
   \STATE {\bfseries Input:} Hub set $\mathcal{H}$, business context descriptions $\mathcal{B}$, optimization model $\mathcal{O}$, iterations $K$
   \STATE {\bfseries Output:} Best capacity solution $\mathbf{l}^* \triangleq \{l_h^*\}_{h \in \mathcal{H}}$
   \STATE Build business context decision table $\mathcal{T}$ from $\mathcal{B}$
   \STATE Propose initial solution $\mathbf{l}^{1}\triangleq\{l_h^{1}\}_{h \in \mathcal{H}} $ based on $\mathcal{T}$
   \STATE $\mathbf{l}^* \leftarrow \mathbf{l}^{1}$
   \FOR{$k = 1$ {\bfseries to} $K$}
       \STATE Evaluate $\mathbf{l}^{k}$ via optimization $\mathcal{O}$
       \STATE Compute routing feedback $\mathcal{R}^{k}$ (e.g., per-hub utilization \%, excess units) from $\mathcal{O}$'s optimal solution
       \IF{$\mathbf{l}^{k}$ respects high-confidence decisions in $\mathcal{T}$ {\bfseries and} routing pattern is reasonable}
           \STATE $\mathbf{l}^* \leftarrow \mathbf{l}^{k}$ \COMMENT{$\Rightarrow$ \emph{Keep}}
       \ELSE
           \STATE Retain $\mathbf{l}^*$ \COMMENT{$\Rightarrow$ \emph{Discard} $\mathbf{l}^{k}$}
       \ENDIF
       \STATE Update decision table $\mathcal{T}$ using $\mathcal{R}^{k}$
       \STATE Propose refined solution $\mathbf{l}^{k+1}$ from $\mathcal{T}$
   \ENDFOR
   \STATE \textbf{return} $\mathbf{l}^*$
\end{algorithmic}
\end{algorithm}

\begin{figure*}[t]
\centering
\includegraphics[width=0.95\textwidth]{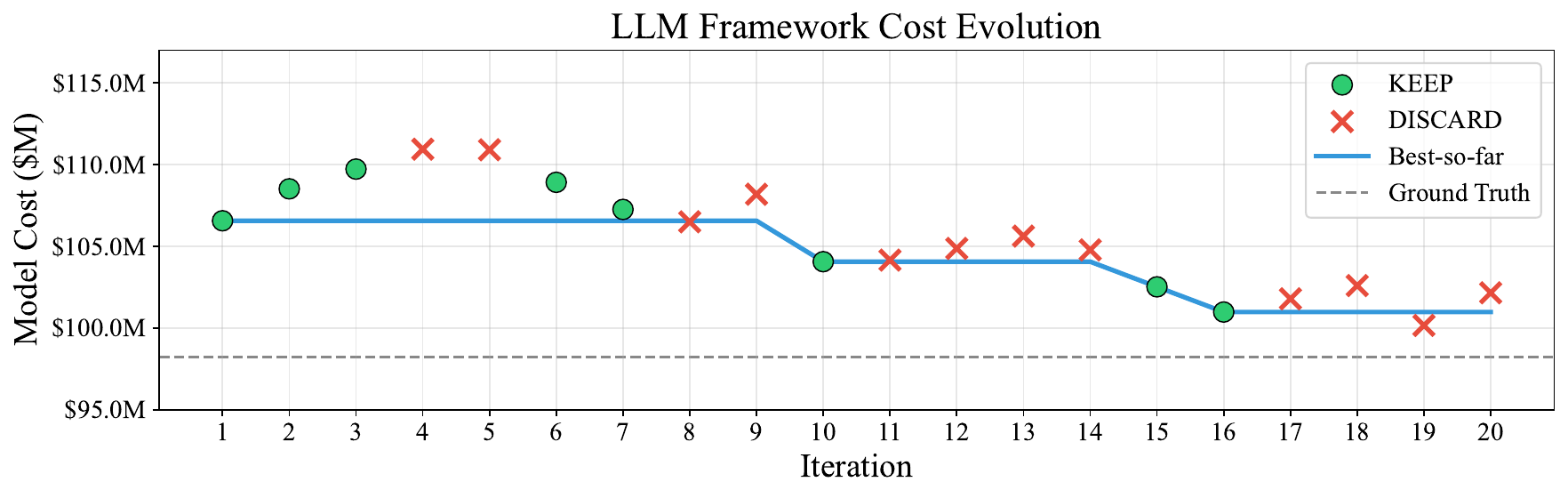}
\caption{Model cost for each iteration's solution, evaluated post-hoc under the true model. The dashed line marks the ground-truth optimal. The best-so-far curve decreases progressively across 20 iterations.}
\label{fig:convergence}
\vspace{-0.1in}
\end{figure*}

\section{Experiments}

\subsection{Setup}
We evaluate our framework on a real-world freight network derived from the Freight Analysis Framework dataset covering the three states (Florida, Georgia, South Carolina) in the southeastern US. The dataset contains 15 qualitative business contexts in natural language that will be incorporate into the optimization model by LLM (See Appendix~\ref{app:context} for details). To measure how well the LLM translates qualitative context into capacity decisions, we construct a controlled evaluation setting.
For each business context, we manually define a corresponding numerical parameter change as the true parameter, which can be one of the following three categories: hub cost multipliers $\alpha_h$ (modifying contracting costs), penalty multipliers $\beta_h$ (modifying overflow penalty rates), and demand multipliers $\gamma_{od}$ (modifying OD-pair demand volumes).
We plug the true parameters into the optimization objective, forming the true optimization model.
The ground-truth solution is obtained by solving the true model, representing the best achievable outcome if every piece of qualitative context were perfectly quantified. The true model will also be used for post-hoc evaluation of the LLM proposed solutions.
Note that we construct the true model and parameters for evaluation purpose only. The LLM agent never observes these numerical values as they are always hidden in reality.

We compare three approaches: 1) \emph{Base Optimal}: Solver-optimum of the optimization model without business context; 2) \emph{LLM Framework}: Solution using the proposed framework with LLM digesting the natural-language context; and 3) \emph{True Optimal}: Solver-optimum of the true optimization model with context-modified parameters. We use Claude Sonnet~4 at temperature~0, run for $K=20$ iterations (see full setup details in Appendix~\ref{app:network}).
We use the optimality gap as the evaluation metric, defined as $(\text{framework solution cost} - \text{True Optimal cost}) / \text{True Optimal cost} \times 100\%$. This metric measures how effectively the framework bridges the gap between textual business inputs and quantitative optimization (see full setup details in Appendix~\ref{app:network}).

\subsection{Results}

\begin{table}[!t]
\centering
\caption{Evaluation of all solutions against the ground truth. Costs are in USD.}
\label{tab:main}
\resizebox{0.7\textwidth}{!}{
\begin{tabular}{@{}lrrrr@{}}
\toprule
Method & Total Cost & Hub Cost & Penalty Cost & Gap (\%) \\
\midrule
\emph{Base Optimal} & \$109.0M & \$44.0M & \$44.1M & 11.0 \\
\emph{LLM Framework} & \$101.0M & \$20.9M & \$59.2M & 2.8 \\
\emph{Ground Truth} & \$98.2M & \$21.2M & \$56.1M & 0.0 \\
\bottomrule
\vspace{-0.35in}
\end{tabular}
}
\end{table}

\paragraph{LLM Closes the Gap}
Table~\ref{tab:main} presents the quantitative results.
The LLM framework achieves 2.8\% optimality gap, a substantial reduce from the 11.0\% gap by base optimal.
The base optimal invests heavily in hubs whose costs are inflated by business context, resulting in \$44.0M in hub costs, more than double the true optimal.
In contrast, the LLM agent correctly identified which expensive hubs to avoid, enabling a hub cost of \$20.9M that is even lower than the \$21.2M in true optimal.
The optimality gap of the LLM framework arises from the slightly higher penalty costs, attributable only to a single hub-level difference, as shown in the hub-level decision comparison in Appendix \ref{app:iterations}.

\paragraph{LLM Provides Right Search Direction}
Figure~\ref{fig:convergence} shows the model cost for each iteration's solution, evaluated post-hoc under the true model. The convergence reveals several notable patterns.
First, the LLM's decisions on whether to keep or discard the solutions align well with the true model objective.
The best-so-far curve (blue) decreases by \$5.6M from iteration~1 to iteration~16, confirming that the LLM's qualitative reasoning produces improvements under the hidden true model.
Second, an interesting discrepancy emerges: iteration~19 that is discarded actually achieves \$100.2M total cost, which is slightly better than the then-current best of \$101.0M.
This near-miss illustrates the inherent challenge of using LLM to optimize a hidden objective that the routing-based LLM heuristic occasionally discards marginally beneficial solutions.
Nevertheless, the overall trajectory demonstrates that chain-of-thought reasoning is a reliable proxy for the true cost structure, closing the gap from 8.5\% at iteration~1 to 2.8\% at iteration~16. In Appendix~\ref{app:casestudies}, we evaluate the LLM's response to isolated qualitative signals through two case studies. Both demonstrate that the LLM correctly interprets single context items and produces meaningful capacity adjustments.

\section{Conclusion}

We presented a framework that couples LLM directional reasoning with two-stage stochastic optimization for hub capacity planning under qualitative business uncertainty.
The LLM agent, given only natural-language context descriptions and routing utilization feedback, achieves a 2.8\% optimality gap against the hidden true optimal. In addition, three key findings emerge:
First, LLMs can perform effective directional cost-benefit reasoning from qualitative descriptions.
Second, iterative routing feedback enables the LLM to discover non-obvious network structure that are difficult to anticipate from context alone. Third, the deliberate separation of routing feedback from cost feedback is essential, especially when business context modifies the cost structure.
Future work includes scaling to larger networks and multiple regions, comparing across LLM architectures, incorporating adaptive prompt refinement based on convergence patterns, and developing theoretical convergence guarantees for the iterative framework.


%


\bibliographystyle{apalike}
\bibliography{references_arxiv}

@book{birge2011introduction,
  author    = {Birge, John R. and Louveaux, Fran{\c{c}}ois},
  title     = {Introduction to Stochastic Programming},
  publisher = {Springer},
  edition   = {2nd},
  year      = {2011}
}

@article{bommasani2021opportunities,
  author  = {Bommasani, Rishi and Hudson, Drew A. and Adeli, Ehsan and others},
  title   = {On the opportunities and risks of foundation models},
  journal = {arXiv preprint arXiv:2108.07258},
  year    = {2021}
}

@inproceedings{wang2025solid,
  author    = {Wang, Yiwen and You, Teng Guo and Boussioux, L{\'e}onard and Liu, Shuai},
  title     = {{SOLID}: A framework of synergizing optimization and {LLMs} for intelligent decision-making},
  booktitle = {NeurIPS ML$\times$OR Workshop},
  year      = {2025}
}

@inproceedings{jovine2025listen,
  author    = {Jovine, Aldo Sandoval and Ye, Tian and Bahk, Francis and Wang, Jiayu and Shmoys, David B. and Frazier, Peter I.},
  title     = {{LISTEN} to your preferences: An {LLM} framework for multi-objective selection},
  booktitle = {NeurIPS ML$\times$OR Workshop},
  year      = {2025}
}

@article{ahmaditeshnizi2024optimus,
  author  = {AhmadiTeshnizi, Ali and Gao, Wenzhi and Udell, Madeleine},
  title   = {{OptiMUS}: Optimization modeling using {MIP} solvers and large language models},
  journal = {arXiv preprint arXiv:2310.06116},
  year    = {2023}
}

@article{liu2024dellma,
  author  = {Liu, Ollie and Khakhar, Jeevan and Yang, Diyi},
  title   = {{DeLLMa}: Decision-making under uncertainty with large language models},
  journal = {arXiv preprint arXiv:2402.02392},
  year    = {2024}
}

@article{heitsch2003scenario,
  author  = {Heitsch, Holger and R{\"o}misch, Werner},
  title   = {Scenario reduction algorithms in stochastic programming},
  journal = {Computational Optimization and Applications},
  volume  = {24},
  number  = {2},
  pages   = {187--206},
  year    = {2003}
}

@inproceedings{wei2022chain,
  author    = {Wei, Jason and Wang, Xuezhi and Schuurmans, Dale and others},
  title     = {Chain-of-thought prompting elicits reasoning in large language models},
  booktitle = {NeurIPS},
  year      = {2022}
}

@inproceedings{yang2024large,
  author    = {Yang, Chengrun and Wang, Xuezhi and Lu, Yifeng and Liu, Hanxiao and Le, Quoc V. and Zhou, Denny and Chen, Xinyun},
  title     = {Large language models as optimizers},
  booktitle = {ICLR},
  year      = {2024}
}

@article{alumur2008network,
  author  = {Alumur, Sibel and Kara, Bahar Y.},
  title   = {Network hub location problems: The state of the art},
  journal = {European Journal of Operational Research},
  volume  = {190},
  number  = {1},
  pages   = {1--21},
  year    = {2008}
}

@article{contreras2011stochastic,
  author  = {Contreras, Ivan and Cordeau, Jean-Fran{\c{c}}ois and Laporte, Gilbert},
  title   = {Stochastic uncapacitated hub location},
  journal = {European Journal of Operational Research},
  volume  = {212},
  number  = {3},
  pages   = {518--528},
  year    = {2011}
}

@inproceedings{yao2024tree,
  author    = {Yao, Shunyu and Yu, Dian and Zhao, Jeffrey and others},
  title     = {Tree of thoughts: Deliberate problem solving with large language models},
  booktitle = {NeurIPS},
  year      = {2024}
}

@article{ye2024reevo,
  author  = {Ye, Shuai and Wang, Yuxin and Zhang, Qingfu},
  title   = {{ReEvo}: Large language models as hyper-heuristics with reflective evolution},
  journal = {arXiv preprint arXiv:2402.01145},
  year    = {2024}
}

@article{hu2021stochastic,
  author  = {Hu, Qian and Hu, Shaolong and Wang, Jian and Li, Xiang},
  title   = {Stochastic single allocation hub location problems with balanced utilization of hub capacities},
  journal = {Transportation Research Part B: Methodological},
  volume  = {153},
  pages   = {1--22},
  year    = {2021}
}

@inproceedings{liu2023logistics,
  author    = {Liu, Xiaoyue and Li, Jingze and Montreuil, Benoit},
  title     = {Logistics hub capacity deployment in hyperconnected transportation network under uncertainty},
  booktitle = {2023 IISE Annual Conference and Expo},
  year      = {2023}
}

@inproceedings{liu2024multi,
  author    = {Liu, Xiaoyue and Li, Jingze and Dahan, Mathieu and Montreuil, Benoit},
  title     = {Multi-period stochastic logistic hub capacity planning for relay transportation},
  booktitle = {2024 IISE Annual Conference and Expo},
  year      = {2024}
}

@inproceedings{liu2025network,
  author    = {Liu, Xiaoyue and Muthukrishnan, Praveen and Montreuil, Benoit},
  title     = {Network design and capacity management in hyperconnected urban logistic networks},
  booktitle = {11th International Physical Internet Conference},
  year      = {2025}
}

@inproceedings{liu2024dynamic,
  author = {Liu, Xiaoyue and Xu, Yujia and Montreuil, Benoit},
  year = {2024},
  title = {Dynamic Containerized Modular Capacity Planning and Resource Allocation in Hyperconnected Supply Chain Ecosystems},
  booktitle = {Proceedings of 10th International Physical Internet Conference (IPIC)}
}

@article{li2023multi,
  title={A multi-agent system for operating hyperconnected freight transportation in the physical internet},
  author={Li, Jingze and Liu, Xiaoyue and Montreuil, Benoit},
  journal={IFAC-PapersOnLine},
  volume={56},
  number={2},
  pages={7585--7590},
  year={2023},
  publisher={Elsevier}
}

@article{li2024stochastic,
  title={Stochastic service network design with different operational patterns for hyperconnected relay transportation},
  author={Li, Jingze and Liu, Xiaoyue and Dahan, Mathieu and Montreuil, Benoit},
  journal={Proceedings of 9th International Physical Internet Conference (IPIC)},
  year={2024}
}

@article{liu2026modular,
  title={Modular and Mobile Capacity Planning for Hyperconnected Supply Chain Networks},
  author={Liu, Xiaoyue and Klibi, Walid and Montreuil, Benoit},
  journal={arXiv preprint arXiv:2601.11107},
  year={2026}
}

@article{liu2026two,
  title={Two-Echelon Delivery Vehicle Sharing and Repositioning in Hyperconnected Urban Logistic Networks},
  author={Liu, Xiaoyue and Montreuil, Benoit},
  journal={arXiv preprint arXiv:2606.17608},
  year={2026}
}

@article{merakli2021robust,
  author  = {Taherkhani, Gita and Alumur, Sibel A. and Hosseini, Mohammadreza},
  title   = {Robust stochastic models for profit-maximizing hub location problems},
  journal = {Transportation Science},
  volume  = {55},
  number  = {6},
  pages   = {1322--1350},
  year    = {2021}
}

@article{liu2025dynamic,
  title={Dynamic hub capacity planning in hyperconnected relay transportation networks under uncertainty},
  author={Liu, Xiaoyue and Li, Jingze and Dahan, Mathieu and Montreuil, Benoit},
  journal={Transportation Research Part E: Logistics and Transportation Review},
  volume={194},
  pages={103940},
  year={2025},
  publisher={Elsevier}
}

@article{huang2025orlm,
  author  = {Huang, Chenyu and Tang, Zhixing and Hu, Shichao and Jiang, Ruoyu and Zheng, Xiaonan and Ge, Dongdong and Wang, Baihua and Wang, Zizhuo},
  title   = {{ORLM}: A customizable framework in training large models for automated optimization modeling},
  journal = {Operations Research},
  year    = {2025}
}

@inproceedings{ramamonjison2023nl4opt,
  author    = {Ramamonjison, Rindranirina and Yu, Tianyu and Li, Raymond and others},
  title     = {{NL4Opt} competition: Formulating optimization problems based on their natural language descriptions},
  booktitle = {NeurIPS 2022 Competition Track},
  pages     = {189--203},
  publisher = {PMLR},
  year      = {2023}
}

@inproceedings{xiao2024chainofexperts,
  author    = {Xiao, Zihao and Zhang, Dongxiang and Wu, Yuze and others},
  title     = {Chain-of-experts: When {LLMs} meet complex operations research problems},
  booktitle = {ICLR},
  year      = {2024}
}

@article{venkatachalam2025integrating,
  author  = {Venkatachalam, Saravanan},
  title   = {Integrating Large Language Models with Network Optimization for Interactive and Explainable Supply Chain Planning: A Real-World Case Study},
  journal = {arXiv preprint arXiv:2508.21622},
  year    = {2025}
}

@article{aghaei2025potential,
  author  = {Aghaei, Reza and Kiaei, Amir Ali and Boush, Mohammadreza and Vahidi, Javad and Barzegar, Zahra and Rofoosheh, Maryam},
  title   = {The potential of large language models in supply chain management},
  journal = {arXiv preprint arXiv:2501.15411},
  year    = {2025}
}

@inproceedings{
dong2025greentea,
title={Green{TEA}: Gradient Descent with Topic-modeling and Evolutionary Auto-prompting},
author={Zheng Dong and Luming Shang and Gabriela Olinto},
booktitle={First International KDD Workshop on Prompt Optimization, 2025},
year={2025},
}

@inproceedings{
dong2023spatiotemporal,
title={Spatio-temporal point processes with deep non-stationary kernels},
author={Zheng Dong and Xiuyuan Cheng and Yao Xie},
booktitle={The Eleventh International Conference on Learning Representations },
year={2023},
}

@article{dong2023non,
  title={Non-stationary spatio-temporal point process modeling for high-resolution COVID-19 data},
  author={Dong, Zheng and Zhu, Shixiang and Xie, Yao and Mateu, Jorge and Rodr{\'\i}guez-Cort{\'e}s, Francisco J},
  journal={Journal of the Royal Statistical Society Series C: Applied Statistics},
  volume={72},
  number={2},
  pages={368--386},
  year={2023},
  publisher={Oxford University Press US}
}

@inproceedings{
zhu2022neural,
title={Neural Spectral Marked Point Processes},
author={Shixiang Zhu and Haoyun Wang and Zheng Dong and Xiuyuan Cheng and Yao Xie},
booktitle={International Conference on Learning Representations},
year={2022},
}

@article{dong2026deep,
  title={Deep Graph Kernel Point Processes over Networks},
  author={Dong, Zheng and Repasky, Matthew and Cheng, Xiuyuan and Xie, Yao},
  journal={Journal of Computational and Graphical Statistics},
  pages={1--34},
  year={2026},
  publisher={Taylor \& Francis}
}

@inproceedings{dong2025conditional,
  title={Conditional generative modeling for high-dimensional marked temporal point processes},
  author={Dong, Zheng and Fan, Zekai and Zhu, Shixiang},
  booktitle={Proceedings of the 31st ACM SIGKDD Conference on Knowledge Discovery and Data Mining V. 1},
  pages={248--259},
  year={2025}
}



\newpage
\appendix

\setcounter{figure}{0} \renewcommand{\thefigure}{\thesection\arabic{figure}}
\setcounter{table}{0} \renewcommand{\thetable}{\thesection\arabic{table}}
\setcounter{equation}{0} \renewcommand{\theequation}{\thesection\arabic{equation}}

\section{Mathematical Formulation}
\label{app:formulation}

Table~\ref{tab:notation} summarizes the notation used in the formulation.

\small
\begin{longtable}{@{}ll@{}}
\caption{Summary of notation.} \label{tab:notation} \\
\toprule
\multicolumn{2}{@{}l}{\emph{Sets}} \\
\midrule
\endfirsthead
\toprule
\multicolumn{2}{@{}l}{\emph{(continued)}} \\
\midrule
\endhead
\bottomrule
\endfoot
$\mathcal{G} = (\mathcal{N}, \mathcal{A})$ & Network graph with node set $\mathcal{N}$ and arc set $\mathcal{A}$ \\
$\mathcal{H} \subseteq \mathcal{N}$ & Set of hub nodes \\
$\mathcal{K} \subseteq \mathcal{N} \times \mathcal{N}$ & Set of origin-destination (OD) pairs \\
$\mathcal{L} = \{0,1,2,3,4\}$ & Set of capacity levels \\
$\Omega$ & Set of demand scenarios \\
$\Phi$ & Set of operational periods \\
\midrule
\multicolumn{2}{@{}l}{\emph{Parameters}} \\
\midrule
$c_h^l$ & Contracting cost for hub $h$ at level $l$ \\
$c^{\text{flow}}$ & Per-unit-time transportation cost (\$/hr) \\
$c^{\text{pen}}_h$ & Penalty cost for excess capacity at hub $h$ (\$/unit) \\
$t_{ij}$ & Travel time on arc $(i,j)$ (hours) \\
$\text{Cap}(l)$ & Weekly throughput capacity at level $l$ (vehicles) \\
$d_{od}^{\omega\phi}$ & Demand for OD pair $(o,d)$ in scenario $\omega$, period $\phi$ \\
$p_\omega$ & Probability of scenario $\omega$ \\
$m$ & Scaling factor: period length / week length \\
\midrule
\multicolumn{2}{@{}l}{\emph{Decision Variables}} \\
\midrule
$x_{h,l} \in \{0,1\}$ & 1 if hub $h$ is assigned level $l$, 0 otherwise \\
$f_{ij}^{od\omega\phi} \geq 0$ & Flow on arc $(i,j)$ for OD pair $(o,d)$, scenario $\omega$, period $\phi$ \\
$e_h^{\omega\phi} \geq 0$ & Excess capacity (overflow) at hub $h$, scenario $\omega$, period $\phi$ \\
\midrule
\multicolumn{2}{@{}l}{\emph{Business Context Multipliers (True Model Only)}} \\
\midrule
$\alpha_h$ & Hub cost multiplier ($\tilde{c}_h^l = \alpha_h \cdot c_h^l$) \\
$\beta_h$ & Penalty cost multiplier ($\tilde{c}^{\text{pen}}_h = \beta_h \cdot c^{\text{pen}}_h$) \\
$\gamma_{od}$ & Demand multiplier ($\tilde{d}_{od}^{\omega\phi} = \gamma_{od} \cdot d_{od}^{\omega\phi}$) \\
\end{longtable}
\normalsize

We formulate the two-stage stochastic hub capacity planning problem on a network $\mathcal{G} = (\mathcal{N}, \mathcal{A})$ with hub nodes $\mathcal{H} \subseteq \mathcal{N}$, OD pairs $\mathcal{K}$, scenarios $\omega \in \Omega$ with probabilities $p_\omega$, and periods $\phi \in \Phi$.
Decision variables: $x_{h,l} \in \{0,1\}$ (hub $h$ assigned level $l$), $f_{ij}^{od\omega\phi} \geq 0$ (flow), $e_h^{\omega\phi} \geq 0$ (excess capacity).
\begin{align}
\min_{x, f, e} \quad & \sum_{h \in \mathcal{H}} \sum_{l \in \mathcal{L}} c_h^l \, x_{h,l} + \sum_{\omega \in \Omega} p_\omega \sum_{\phi \in \Phi} \left[ \sum_{(i,j) \in \mathcal{A}} c^{\text{flow}} \, t_{ij} \sum_{(o,d) \in \mathcal{K}} f_{ij}^{od\omega\phi} + \sum_{h \in \mathcal{H}} c^{\text{pen}}_h \, e_{h}^{\omega\phi} \right] \label{eq:obj} \\
\text{s.t.} \quad & \sum_{l \in \mathcal{L}} x_{h,l} = 1, \quad \forall h \in \mathcal{H} \label{eq:one_level} \\
& \sum_{j:(o,j) \in \mathcal{A}} f_{oj}^{od\omega\phi} = d_{od}^{\omega\phi}, \quad \forall (o,d), \omega, \phi \label{eq:origin} \\
& \sum_{i:(i,d) \in \mathcal{A}} f_{id}^{od\omega\phi} = d_{od}^{\omega\phi}, \quad \forall (o,d), \omega, \phi \label{eq:dest} \\
& \sum_{j:(h,j)} f_{hj}^{od\omega\phi} = \sum_{i:(i,h)} f_{ih}^{od\omega\phi}, \quad \forall h \in \mathcal{H}, (o,d), \omega, \phi \label{eq:conserv} \\
& \sum_{(o,d) \in \mathcal{K}} \sum_{i:(i,h)} f_{ih}^{od\omega\phi} \leq m \left( \sum_{l \in \mathcal{L}} \text{Cap}(l) \, x_{h,l} + e_h^{\omega\phi} \right), \quad \forall h, \omega, \phi \label{eq:cap} \\
& x_{h,l} \in \{0,1\}, \quad f \geq 0, \quad e \geq 0 \nonumber
\end{align}

where $c_h^l$ is the contracting cost for hub $h$ at level $l$, $c^{\text{flow}}$ is the per-unit-time transportation cost, $t_{ij}$ is the travel time on arc $(i,j)$, $c^{\text{pen}}_h$ is the penalty cost for excess capacity at hub $h$, $d_{od}^{\omega\phi}$ is the demand for OD pair $(o,d)$ in scenario $\omega$ and period $\phi$, and $m$ is a scaling factor converting weekly capacity to the operational period length.
Constraint~\eqref{eq:one_level} ensures exactly one level per hub; \eqref{eq:origin}--\eqref{eq:dest} enforce demand satisfaction; \eqref{eq:conserv} is flow conservation at hubs; and \eqref{eq:cap} links capacity decisions to flow with excess variables.

The true model modifies these parameters with business context multipliers: $\tilde{c}_h^l = \alpha_h \cdot c_h^l$, $\tilde{c}^{\text{pen}}_h = \beta_h \cdot c^{\text{pen}}_h$, and $\tilde{d}_{od}^{\omega\phi} = \gamma_{od} \cdot d_{od}^{\omega\phi}$.

\section{Business Context Descriptions}
\label{app:context}

The following 15 natural-language descriptions are provided to the LLM agent. The numerical multipliers in parentheses represent the manually-constructed true parameters. The LLM never sees the numerical multipliers; these are shown here for reference only.

\begin{enumerate}[itemsep=2pt, parsep=2pt, topsep=2pt]
\small
    \item Hub 52 (Savannah area) is facing new EPA environmental compliance requirements due to its proximity to wetlands. Permitting and mitigation costs have increased its operating expenses roughly ninefold. \emph{($\alpha_{52} = 9$)}
    \item Hub 42 (central Florida) has seen its insurance premiums skyrocket after a series of cargo theft incidents and a warehouse fire. Operating costs have increased roughly sevenfold. \emph{($\alpha_{42} = 7$)}
    \item Hub 64 (South Carolina) is facing a severe labor shortage driven by competition from new facilities. Wage increases have pushed operating costs to roughly six times their previous level. \emph{($\alpha_{64} = 6$)}
    \item Hub 44 (north Florida) has been cited for structural deficiencies. Emergency repairs have roughly quintupled its operating costs. \emph{($\alpha_{44} = 5$)}
    \item Hub 47 (central Georgia) has negotiated a long-term contract with a new carrier consortium, cutting its operating costs by about 70\%. \emph{($\alpha_{47} = 0.3$)}
    \item Hub 48 (south Georgia) has completed a major automation upgrade, cutting its operating costs by about 65\%. \emph{($\alpha_{48} = 0.35$)}
    \item Hub 49 (Atlanta) has received a state economic development incentive, cutting its effective operating costs by about 70\%. \emph{($\alpha_{49} = 0.3$)}
    \item Hub 43 (Jacksonville) recently completed major infrastructure upgrades, reducing its operational costs by about half. \emph{($\alpha_{43} = 0.5$)}
    \item The Savannah port (near hub 52) is experiencing severe congestion. Overflow penalties at hub 52 are now roughly four times normal rates. \emph{($\beta_{52} = 4$)}
    \item A severe hurricane season is expected in southern Florida. Overflow penalties at hubs 41, 42, and 45 are now roughly triple normal rates. \emph{($\beta_{41,42,45} = 3$)}
    \item Hub 64 workers are threatening a strike. Overflow penalties at hub 64 are now roughly four times normal. \emph{($\beta_{64} = 4$)}
    \item The greater Atlanta metropolitan area is experiencing a logistics boom. Freight demand in the region has grown by about 50\%. \emph{($\gamma_{\text{Atlanta}} = 1.5$)}
    \item The Charleston region is seeing rapid e-commerce growth. Freight volumes have grown by about 50\%. \emph{($\gamma_{\text{Charleston}} = 1.5$)}
    \item Freight demand from Jacksonville to Miami has surged due to a port expansion---volumes have roughly doubled. \emph{($\gamma_{\text{Jax-Mia}} = 2.0$)}
    \item Hub 51 (east Georgia) has been designated a state opportunity zone, reducing its operating costs by about half. \emph{($\alpha_{51} = 0.5$)}
\end{enumerate}

\section{Prompt Template}
\label{app:prompt}

The LLM prompt follows a structured template with six components:

\begin{enumerate}[itemsep=2pt, parsep=2pt, topsep=2pt]
    \item \textbf{Problem description}: Defines the two-stage stochastic hub capacity planning problem, including the objective (minimize total cost = hub contracting + expected transportation + expected penalty).
    \item \textbf{Network data}: Hub IDs, OD pairs, arc structure, and geographic context.
    \item \textbf{Cost structure}: Per-hub contracting costs at each capacity level, transportation cost per unit-time, and base penalty cost.
    \item \textbf{Capacity levels}: Mapping from levels 0--4 to weekly throughput $\{0, 30, 60, 90, 120\}$ vehicles.
    \item \textbf{Business context}: The natural-language descriptions (Appendix~\ref{app:context}).
    \item \textbf{Output format}: JSON mapping each hub ID to a capacity level, with keep/discard decision and reasoning for iterations $\geq 2$.
\end{enumerate}

The keep/discard guidelines instruct the agent to decide on the basis of directional reasoning and routing evidence, \emph{not} on the optimization model's reported cost. Although the routing oracle prints cost figures, the agent is explicitly told these reflect the original (base) parameters, do not capture business context effects, and must be ignored for decision-making. The criteria are: (i)~\emph{Keep} a solution if it respects all high-confidence business context decisions and its routing patterns are reasonable (capacity is well matched to observed demand, with no unexpected bottlenecks or wasted capacity); (ii)~\emph{Discard} and revert to the current best otherwise. High-confidence context decisions are held firm even when the base-model cost rises, whereas low-confidence decisions are allowed to follow the routing-implied optimal.

\section{Detailed Experimental Setup}
\label{app:network}

\textbf{Network.}
We use a freight relay transportation network derived from the Freight Analysis Framework dataset (www.bts.gov/faf) covering the southeastern United States.
The 3-state region (Florida, Georgia, South Carolina) contains 13~hub nodes, 22~origin/destination nodes, and multiple OD pairs connected by directed arcs with travel times. Table~\ref{tab:network} summarizes the network characteristics for the 3-state region used in our experiments.

\textbf{Scenarios.}
Demand scenarios are generated via a Poisson process from yearly OD-pair flow volumes: for each OD pair with yearly flow $\lambda > 0$, we generate 50~raw scenarios using $\text{Poisson}(\lambda / 12)$ for each of 12~monthly periods.
These are reduced to 10~representative scenarios via Fast Forward Selection (FFS)~\citep{heitsch2003scenario} with associated probabilities.
For computational tractability in the final evaluation, we use 3~scenarios and 4~periods.

\textbf{Business context.}
The LLM receives the natural-language descriptions as input, e.g., ``Hub~52 is facing new EPA environmental compliance requirements; operating expenses have increased roughly ninefold''.
The true model incorporates the context items via true multipliers, including:
hub cost multipliers ranging from $5\times$ to $9\times$ increases (environmental compliance, insurance, labor shortages, structural repairs) and $50\%$--$70\%$ decreases (automation, incentives, infrastructure upgrades);
penalty multipliers of $3\times$--$4\times$ (port congestion, hurricane risk, strike threats);
and demand shifts of $+50\%$ to $+100\%$ (logistics booms, port expansions).




\begin{table}[h]
\centering
\caption{Network statistics for the 3-state (FL, GA, SC) region.}
\label{tab:network}
\small
\begin{tabular}{@{}lr@{}}
\toprule
Attribute & Value \\
\midrule
Hub nodes $|\mathcal{H}|$ & 13 \\
Origin/destination nodes & 9 \\
Total nodes $|\mathcal{N}|$ & 22 \\
Directed arcs $|\mathcal{A}|$ & 436 \\
OD pairs $|\mathcal{K}|$ & 72 \\
Capacity levels $|\mathcal{L}|$ & 5 (0--4) \\
Raw demand scenarios & 50 \\
Reduced scenarios $|\Omega|$ & 3 \\
Operational periods $|\Phi|$ & 4 \\
Base transportation cost $c^{\text{flow}}$ & \$20/hr per unit \\
Base penalty cost $c^{\text{pen}}$ & \$2{,}000/unit (scaled) \\
\bottomrule
\end{tabular}
\end{table}

\section{Detailed Main Results}
\label{app:iterations}

\subsection{Hub-Level Decision Comparison}
\label{app:hub_level_dec}

\begin{table}[t]
\centering
\caption{Hub-level decisions across methods. Bold indicates a difference from the true optimal.}
\label{tab:levels}
\small
\begin{tabular}{@{}lccccccccccccc@{}}
\toprule
Method & 41 & 42 & 43 & 44 & 45 & 46 & 47 & 48 & 49 & 50 & 51 & 52 & 64 \\
\midrule
\emph{Base Optimal} & 0 & \textbf{4} & 4 & \textbf{2} & 0 & 0 & \textbf{1} & 2 & 0 & 0 & 1 & 4 & 4 \\
\emph{LLM Framework} & 0 & 0 & 4 & 0 & 0 & 0 & 2 & \textbf{1} & 0 & 0 & 1 & 4 & 4 \\
\emph{True Optimal} & 0 & 0 & 4 & 0 & 0 & 0 & 2 & 4 & 0 & 0 & 1 & 4 & 4 \\
\bottomrule
\end{tabular}
\end{table}

Table~\ref{tab:levels} compares the hub-level decisions across all three methods.
The LLM framework solution matches the true optimal on 12 of 13 hubs.
Both correctly set five hubs to level~0 (dead hubs that attract no traffic) and both maximize capacity at hubs~43, 52, and 64.
The only difference is hub~48, where it is set to level~1 in LLM framework solution and level~4 in the true optimal.
The base optimal, unaware of cost multipliers, over-invests at hubs~42 and~44 while under-investing at hub~47.

\subsection{Selected Iteration Decision Tables}
We present the LLM's directional reasoning at three key iterations to illustrate the decision-making process.

\paragraph{Iteration 1 (Initial proposal).}
The LLM builds its first decision table entirely from business context, before receiving any routing feedback.
Representative entries:

\begin{quote}
\small
\emph{Hub 52}: Context says $9\times$ cost increase + $4\times$ penalty. Direction: invest \emph{less}. But hub 52 has the highest baseline demand (1170 units). Decision: Level~2 (moderate investment despite high cost). Confidence: \textsc{High}.

\emph{Hub 47}: Context says 70\% cost reduction. Direction: invest \emph{more}. At 30\% of base cost, even level~4 is very cheap (\$382K effective). Decision: Level~4. Confidence: \textsc{High}.

\emph{Hub 49}: Context says 70\% cost reduction + Atlanta demand $+50\%$. Direction: invest \emph{more}. Decision: Level~4. Confidence: \textsc{High}.
\end{quote}

After evaluation, the LLM discovers that hub~49 attracts \emph{zero} traffic despite level~4 capacity---the first instance of geographic locking.

\paragraph{Iteration 10 (Cost-multiplier breakthrough).}
The LLM recalculates whether expensive hubs are worth any investment:

\begin{quote}
\small
\emph{Hub 42} ($7\times$ cost, $3\times$ penalty): At level~1, hub cost = \$609K $\times$ 7 = \$4.26M. Penalty savings = 120 units $\times$ \$25.7K = \$3.09M. Net cost of level~1 = \$1.17M \emph{loss}. Decision: Level~0 is cheaper than level~1. Confidence: \textsc{High}.

\emph{Hub 44} ($5\times$ cost): At level~1, hub cost = \$609K $\times$ 5 = \$3.05M. Penalty at level~0 = 93 units $\times$ \$8.6K = \$0.80M. Level~0 saves \$2.25M. Decision: Level~0. Confidence: \textsc{High}.
\end{quote}

This iteration produces the largest single improvement, saving \$2.78M in estimated context-adjusted cost.

\paragraph{Iteration 16 (Cascade exploitation).}
The LLM maximizes hub~64 after discovering the cascade effect:

\begin{quote}
\small
\emph{Hub 64} ($6\times$ cost): Level~3 $\rightarrow$ Level~4 costs \$1.93M more. But hub~52 excess drops by 120 units $\times$ \$34.3K ($4\times$ penalty) = \$4.11M saved. Net savings = \$2.18M. Each level increase at hub~64 has consistently produced this $4\times$ routing amplification across iterations 7, 15, and 16. Decision: Level~4. Confidence: \textsc{High}.
\end{quote}

This is the final kept solution and the best overall.

\section{Case Study Detailed Results}
\label{app:casestudies}

We evaluate the LLM's ability to respond to isolated qualitative signals through two case studies, each with a single business context item and a 10-iteration loop.
All solutions are compared against a base reference with all hubs at level~0 and base model cost \$37.3M.
Table~\ref{tab:case} summarizes both case studies.

\begin{table}[h]
\centering
\caption{Case study summary. Each uses a single business context item with 10 iterations.}
\label{tab:case}
\small
\begin{tabular}{@{}llrr@{}}
\toprule
Case & Context & Best Cost (\$M) & vs.\ Base \\
\midrule
CS1 & Hub 52 shutdown & 29.3 & $-$21.5\% \\
CS2 & Savannah demand $3\times$ & 29.2 & $-$21.9\% \\
\bottomrule
\end{tabular}
\end{table}

\paragraph{Case Study 1: Hub 52 shutdown.}
Context: ``Hub~52 provider is shutting down operations next year. Do not contract with this hub.''
The LLM correctly interprets this as a hard constraint ($l_{52} = 0$) and compensates by maximizing hub~42 (which has no cost multiplier in this single-context scenario) to absorb Florida traffic, and maintaining hub~64 at level~4 for the SC/GA corridor.
The LLM discovers that hub~52's traffic is largely geographically locked---most of it becomes penalty rather than rerouting to Georgia hubs.
Table~\ref{tab:cs1} shows the iteration trajectory.

\begin{table}[h]
\centering
\caption{Case Study 1 iteration history. Hub 52 is fixed at level~0 throughout.}
\label{tab:cs1}
\small
\begin{tabular}{@{}clrl@{}}
\toprule
Iter & Key changes from previous & Cost (\$M) & Decision \\
\midrule
1 & Main best minus hub 52 & 31.4 & Keep \\
2 & Max Georgia hubs (47/48/51 at L4) & 33.9 & Discard \\
3 & Hub 47 at L3 & 31.7 & Discard \\
4 & Reconfirm iter 1 & 31.4 & Keep \\
5 & Hub 51 at L2 & 31.7 & Discard \\
6 & Add hubs 42 L1, 44 L1 & 30.5 & Keep \\
7 & Hub 42 at L2 & 30.1 & Keep \\
8 & Hub 42 at L4 & \textbf{29.3} & Keep \\
9 & Also increase 44/47/48/51 & 29.9 & Discard \\
10 & Hub 48 at L2 only & 29.6 & Discard \\
\bottomrule
\end{tabular}
\end{table}

The LLM correctly identifies that without hub~52, hub~42 (which has no cost multiplier in this single-context scenario) becomes the primary candidate for absorbing Savannah-corridor traffic.
It progressively increases hub~42 from level~0 to level~4, discovering the same routing amplification effect observed in the main experiment.

\paragraph{Case Study 2: Savannah demand triples.}
Context: ``Freight demand from the Savannah region is expected to triple due to a new fulfillment center.''
The LLM increases capacity at nearby Georgia hubs (47 to level~3, 51 to level~2) as cheap buffer capacity for the anticipated demand surge, while maintaining the main-loop best solution's structure.
Notably, the LLM reasons that even though the optimization model cannot show the demand increase effect (it uses fixed scenarios), positioning spare capacity at low-cost hubs near Savannah is the correct directional response.
Table~\ref{tab:cs2} shows the trajectory.

\begin{table}[h]
\centering
\caption{Case Study 2 iteration history. Savannah demand expected to triple.}
\label{tab:cs2}
\small
\begin{tabular}{@{}clrl@{}}
\toprule
Iter & Key changes from previous & Cost (\$M) & Decision \\
\midrule
1 & Main loop best solution & 28.5 & Keep \\
2 & Max Georgia hubs (47/48/51 at L4) & 31.1 & Discard \\
3 & Hub 47 L3, 48 L2, 51 L2 & 29.5 & Keep \\
4 & Hub 51 at L2 only & 28.9 & Keep \\
5 & Hub 51 L3, 47 L3, 48 L2 & 29.8 & Discard \\
6 & Hub 47 L4, 48 L2, 51 L3 & 30.1 & Discard \\
7 & Reconfirm main best & 28.5 & Keep \\
8 & Hub 47 L3, 51 L2 & \textbf{29.2} & Keep \\
9 & Hub 47 L3, 48 L2 & 29.2 & Keep \\
10 & Hub 47 L4, 51 L2 & 29.5 & Discard \\
\bottomrule
\end{tabular}
\end{table}

The LLM's reasoning is notable: it acknowledges that the optimization model's fixed scenarios cannot reflect the demand tripling, yet it proactively positions spare capacity at low-cost Georgia hubs (47 at 30\% cost, 51 at 50\% cost) near Savannah as a buffer.
The best solution (iteration~8) adds moderate spare capacity at hubs~47 and~51 rather than over-investing, reflecting the LLM's uncertainty about the exact magnitude of the demand increase.


\end{document}